\documentclass[sigconf]{acmart}
\AtBeginDocument{%
  }

\setcopyright{acmlicensed}
\copyrightyear{2018}
\acmYear{2018}
\acmDOI{XXXXXXX.XXXXXXX}
\acmConference[Conference acronym 'XX]{Make sure to enter the correct
  conference title from your rights confirmation email}{June 03--05,
  2018}{Woodstock, NY}
\acmISBN{978-1-4503-XXXX-X/2018/06}

\usepackage{subcaption}

\begin{document}

\title{GFM4GA: Graph Foundation Model for Group Anomaly Detection}

\author{Jiujiu Chen}
\email{jchen349@connect.hkust-gz.edu.cn}
\affiliation{
  \institution{HKUST(GZ)}
  \country{China}
}

\author{Weijun Zeng}
\email{messizeng@tencent.com}
\affiliation{
  \institution{Tencent}
  \country{China}
}

\author{Shaofeng Hu}
\email{hugohu@tencent.com}
\affiliation{
  \institution{Tencent}
  \country{China}
}

\author{Sihong Xie}
\authornote{Sihong Xie is the corresponding author.}
\email{sihongxie@hkust-gz.edu.cn}
\affiliation{
  \institution{HKUST(GZ)}
  \country{China}
}

\author{Hui Xiong}
\email{xionghui@ust.hk}
\affiliation{
  \institution{HKUST}
  \country{Hong Kong, SAR, China}
}

\renewcommand{\shortauthors}{}

\begin{abstract}
Group anomaly detection is crucial in many network applications, but faces challenges due to diverse anomaly patterns. Motivated by the success of large language models (LLMs) in natural language processing, graph foundation models (GFMs) is proposed to handle few-shot learning task with fewer labeling efforts. GFMs have been successfully applied to detection of individual anomalies but cannot be generalized to group anomalies, as group anomaly patterns must be detected as a whole and individuals in an abnormal group can look rather normal. Therefore, we propose GFM4GA, a novel graph foundation model for group anomaly detection. The pipeline is pretrained via dual-level contrastive learning based on feature-based estimation and group extraction, to capture potential group anomaly structure and feature inconsistencies. In the downstream tasks, the pipeline is finetuned in parameter-constrained and group-anomaly-proportion weighted few-shot settings, and its adaptive ability to unseen group anomalies expanded via group contexts determined by labeled anomaly neighbors. Experiments show that GFM4GA surpasses group anomaly detectors and GFMs for individual anomalies, achieving average improvements of 2.85\% in AUROC and 2.55\% in AUPRC.
\end{abstract}




\maketitle

\section{Introduction}
\begin{figure}[htbp]
    \centering
    \includegraphics[width=\linewidth]{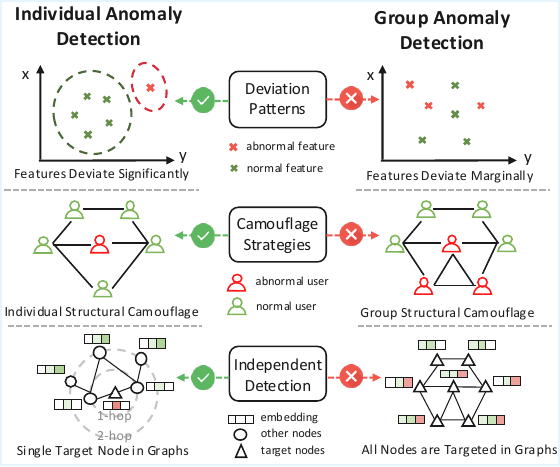}
    \caption{Challenges and difference between individual and group anomaly detections.}
    \label{fig:challenge}
\end{figure}

Graph anomaly detection (GAD), a vital task of graph machine learning, has wide applications across social, co-review, financial, and numerous other graph domains. Graph anomalies are deviations from the expected graph patterns, including feature anomalies~\cite{Zou0L24} that indicates unusual attributes of nodes, and structure anomalies~\cite{ZhaoJSJ22} that highlights irregular connections in the graph. General GAD methods are good at specific anomalies, but they still have limitations in transferability and scalability unless labeled data are sufficient. Recently, Graph Foundation Models (GFMs) show remarkable success of adaptability in multiple graph tasks~\cite{XiaK24, MaoT24}. Once pretrained on massive graph corpora, GFMs can be seamlessly transferred to novel downstream tasks with few-shot adaptions, a capability that graph anomaly detection urgently demands, especially when labeled instances are extremely scarce for novel types of anomaly. 

\begin{figure}[htbp]
    \centering
    \includegraphics[width=\linewidth]{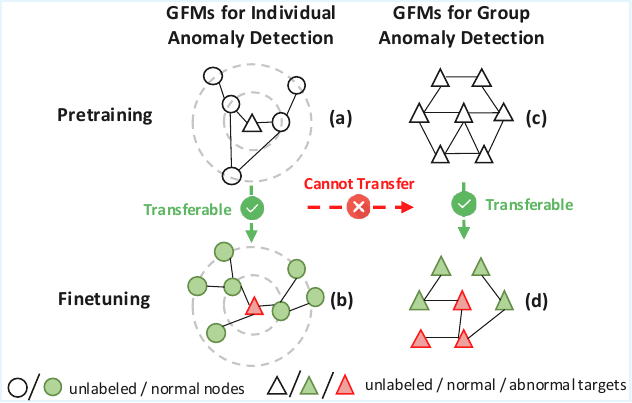}
    \caption{Individual anomalies in (a) and (b) on the left can be transferable, and group anomalies in (c) and (d) on the right can be transferable as well. However, GFM for individual anomaly cannot or hardly transfer to group anomalies via few-shot adaptation or finetuning.}
    \label{fig:toy}
\end{figure}
\begin{figure*}
    \centering
    \includegraphics[width=1.02\linewidth]{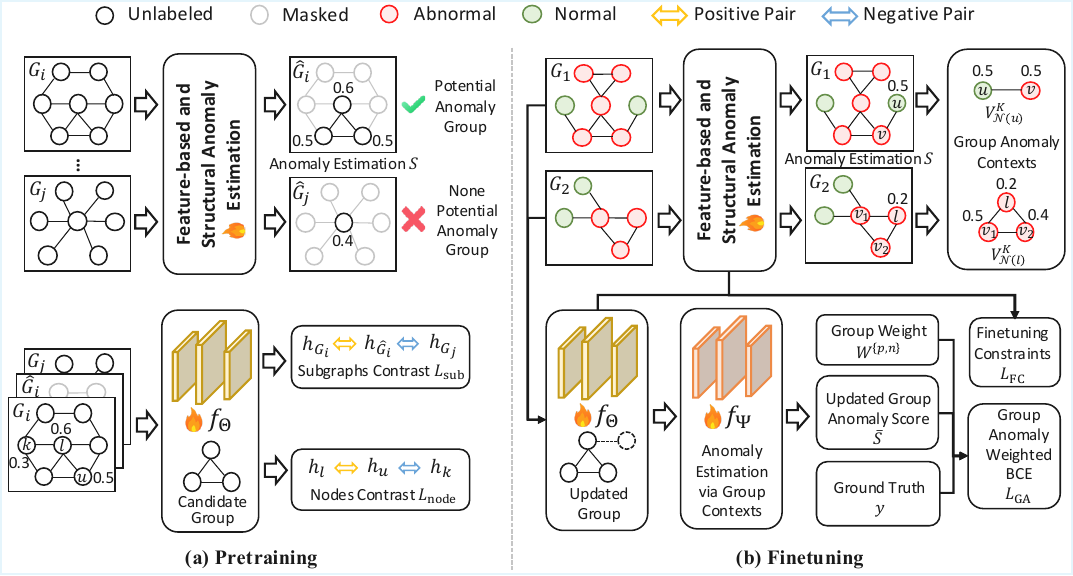}
    \caption{The framework of GFM4GA for group anomaly detection. }
    \label{fig:pipeline}
\end{figure*}

In real world graph applications, abnormal phenomena usually occur in groups, such as anomalies in the right side of Figure~\ref{fig:challenge}, which is defined as group anomaly~\cite{YuHL15}. Anomalies in groups usually have similar patterns, and provide evidences or explanations for each other in crime or anomaly analysis, especially when individual seems normal~\cite{FerozeDAH21}. For example, abnormal group reviewers may play similar roles to propagate or comment similar contents on the goods to attract customers, where abnormal individual reviewer is hard to detect without the group context. Traditional group anomaly detection methods are mainly categorized into community aware and structural deviation approaches. Researches~\cite{AraujoPGFBSPK14, YuHL15, EswaranF18, ZhangZ22, Luo22, LiCSG23} treat group anomaly as communities with low cohesion, and other works~\cite{EberleH07, MukherjeeLG12, BeutelXGPF13, Hooi16} address the abnormal groups from the view of unusual graph structures. Those methods focus on specific types of group anomaly, and they cannot learn common knowledge about group anomalies and quickly adapt to novel ones. 

Although certain GFMs excel at detecting individual anomaly, they encounter unique challenges in group anomaly detection as mentioned in Figure~\ref{fig:challenge}. The \textbf{first challenge} is the marginal deviation of the features, which cannot construct discriminative class prototypes, so individuals in group anomalies are hard to detect without group context. The \textbf{second challenge} lies in the complexity of group structural camouflage, where members of group anomalies have both dense inner connection with each other and inter connection with normal users, reducing the distinguishable residual information and structural patterns. The \textbf{third challenge} is the difference in detecting object. In individual anomaly detectors, the detecting objects are target nodes, and each target node has the supplementary information of an induced subgraph extracted from its neighbors, where the fixed penalty weights for abnormal nodes can be easily applied in the optimization. However, the detecting object of group anomaly is the given subgraph. We cannot directly apply standard binary cross-entropy (BCE) loss or simply increase the penalty of minority in the optimization, because there is no fixed size of group anomalies in the subgraphs, and we cannot know whether group anomalies account for majority or minority in the subgraphs. Besides, we also need consider potential differences in group patterns. For instance, the group anomaly in (c) of Figure~\ref{fig:toy} typically occurs in social or co-review networks, where similar comments are accompanied by dense interconnections. In contrast, the group anomaly depicted in (d) of  Figure~\ref{fig:toy} is common in financial networks or cybernetworks, characterized by a linear structure with as few connections as possible. Those challenges hinder the generalization of GFM for individual anomaly to group anomaly detection. Therefore, it is necessary to build an end-to-end GFM for group anomaly detection. 

To address above challenges, we introduce GFM4GA, a graph foundation model target for group anomaly detection as Figure~\ref{fig:pipeline} shows. \textbf{Firstly}, to extract potential group anomaly part in the given subgraphs, we introduce a feature-based anomaly estimation module and exploit subgraph-subgraph and node-node contrastive learning tasks to capture the structural patterns and feature inconsistencies of group anomaly. \textbf{Secondly}, we select neighbors combined with high degrees and anomaly probabilities to construct group anomaly context for each node in the subgraph, providing key discriminative information to weaken the influence of group structural camouflage. \textbf{Thirdly}, a parameter-constrained and group anomaly weighted finetuning strategy is utilized to focus on the detection object in subgraphs with diverse sizes and group anomalies. The constraints on pretrained parameters keep the common group anomaly knowledge, and weighted few-shot finetuning quickly adapt the pretrained model to novel group anomalies. We experimentally evaluate the proposed pipeline in group anomaly detection tasks across graph domains in few-shot settings. 

\section{Related Work}
We survey recent approaches from the perspective of general GAD and GFM for GAD methods. \\
\textbf{General GAD}. Traditional graph anomaly detection methods relied on graph-centric features such as node degrees, eigenvectors, and centralities et al, which lacks of learnable knowledge that can be transferred into other anomalies~\cite{AkogluTK15}. Recently, general-purpose deep-learning approaches to graph anomaly detection have predominantly relied on contrastive learning and reconstruction-based objectives~\cite{QiaoTAKAP25}. Different unsupervised or self-supervised contrastive learning tasks such as node-level~\cite{JinLZCLP21}, subgraph-level~\cite{ZhengJLCPC21}, node-subgraph~\cite{LiuLPGZK21, YuLCH0W25} and hybrid multi-view~\cite{DuanW0ZHJ0D23, HuTNNLS2023} contrasts were employed in GNN based graph anomaly detection methods. To avoid oversmoothing issues in GNNs, GADAM~\cite{ChenZY024} implemented neighbor-free attention strategies to estimate anomaly. Due to the obvious deviations between reconstruction from hidden aggregated representations and initial patterns, feature-structure reconstruction approach is proposed to discover anomaly patterns~\cite{RoySLYES024, HeXJWH24}.\\
In the scope of group anomaly detection, GLAD~\cite{YuHL15} proposed a hierarchical Bayes model with pairwise and point-wise social media data. Research~\cite{MukherjeeLG12} introduced group spamming indicators in review network, with a relation-based approach to rank candidate groups. Recently, AS-GAE~\cite{ZhangZ22} extracted possible anomalous subgraph from the mismatch of reconstructions, and ComGA~\cite{Luo22} detect group anomaly via community-aware reconstructions. However, the concept of community assumes dense connections inside the community and sparse connections beyond, which is different from our group structural camouflage. \\
\textbf{GFMs for GAD.} Generally, GFMs can be categorized into LLM-based and GNN-based models~\cite{LiuYS25}. LLM-based GFMs unify graph tasks with textual descriptions on node features and graph structures, and have good performance on text-attributed graphs~\cite{YeZWXZ24}. However, LLM modules in the LLM-based GFMs still handle textual sequences, and need further finetuning or alignment training to understand continuous features~\cite{WangFHTHT23, ChenMLJWWWYFLT23}. Besides, the max length of input tokens in LLMs limits LLM-based GFMs to handle large graphs. Recently, GNN-based GFMs for graph anomaly detection have been explored, especially improving the generalization capability of GAD models through transfer learning based on foundation models~\cite{Pan2025survey}. GUDI~\cite{Li024} followed a GFM pipeline to incorporate a self-supervised pretraining approach to capture general patterns across graph domains, and designed few-shot-oriented adaption to specific individual anomaly. ARC~\cite{LZCZP24} proposed a kind of "one-for-all" GFM for capturing universal graph anomaly patterns, which can be transferred with few-shot in-context learning. Motivated by graph prompt methods in conventional graph tasks, UNPrompt~\cite{Niu24} proposed to learn generalized neighborhood prompts to capture latent node attributes for normal and abnormal nodes across graph datasets in zero-shot scenarios, AnomalyGFM~\cite{Qiao25} introduced a GAD-oriented graph foundation model that supports zero-shot inference and few-shot prompt tuning in diverse graph datasets. Those GFMs for graph anomaly detection focused on individual anomaly, which cannot address the issues mentioned in group anomaly detection task. Therefore, it is necessary to propose a GFM targeted for the challenges in group anomaly detection.

\section{Preliminary}
\textbf{Notations.} Given the graph dataset $\mathcal{G}=\{G_1, G_2, ..., G_M\}$, where each ${G_i}=(V_i,E_i,X_i,R_i)$ is a directed or undirected subgraph, with node set $V_i$, edge set $E_i$, node features $X_i \in \mathbb{R}^{n_i \times d}$, $n_i$ denotes node number, $d$ is the dimension of node features, and edge types set $|{R}_i| \ge 1$. The node sets $\{V_1, V_2, ..., V_M\}$ only contain users, and each edge type set in $\{R_1, R_2, ..., R_M\}$ is a subset of the full edge types set $\mathcal{R} = \{R_1 \cup R_2 ... \cup R_M \}$ and $|R_i| \le |\mathcal{R}|$. Only a few subgraphs in $\mathcal{G}$ are manually labeled, which denotes as $\mathcal{G}_L$, and the remaining subgraphs are unlabeled, which is $\mathcal{G}_U$.\\
\textbf{Problem Setting.} Our goal is to pretrain a GFM for group anomaly detection $f_\Theta$ on massive unlabeled subgraphs $\mathcal{G}_U$ with anomaly-related pre-text tasks. Then, k-shot labeled subgraphs $\mathcal{G}_L^k$ randomly selected from $\mathcal{G}_L$ are used to finetune the GFM to adapt to specific group anomaly, and verify its performance on the remaining labeled subgraphs $\mathcal{G}_L/\mathcal{G}_L^k$. Each node $v$ in $\mathcal{G}_L$ has the ground-truth label $y_v \in \{0, 1\}$, where $y_v = 1$ indicates the node is a member of group anomaly, and $y_v = 0$ denotes a benign user or victim. The size of the actual group anomaly varies from subgraphs and graph domains.

\section{Methodology}
\subsection{Feature-based Estimation and Group Extraction}
To capture potential group deviations, the feature-based anomaly estimation with principal component analysis (PCA)~\cite{Mack1993PCA} module $f_\Phi \circ f_\text{PCA}$ is firstly employed to approximate anomaly probability for each node across all subgraphs in the dataset $\mathcal{G}$:
\begin{equation}
    S_i = f_\Phi \circ f_\text{PCA}(X_i)
\end{equation}
where $S_i \in \mathbb{R}^{|V_i|}$ indicates individual anomaly probability of nodes in subgraph $G_i$. Those connected nodes with high anomaly probability can be recognized as a group, which is a smaller subgraph of initial subgraph $\hat{G}_i=\{\hat{V}_i,\hat{E}_i, \hat{X}_i, \hat{R}_i\} \subseteq G_i$. Motivated by Fraudar~\cite{Hooi16}, the node subset $\hat{V}_i$ is determined by the average anomaly probabilities of connected nodes in $G_i$:
\begin{equation}
     \hat{V}_i =\arg \max_{k} \hat{S}_i^k = \arg \max_{k} \frac{\sum_{k=1}^{K^{*}}{S_i^k}}{K^*},
\end{equation}
where the denominator $K^*$ indicates the optimal number of extracted nodes, and the numerator denotes the sum of extracted node anomaly probabilities. Then, the corresponding edge set $\hat{E}_i$, node features $\hat{X}_i$ and relation set $\hat{R}_i$ are determined by the extracted node set $\hat{V}_i$. Here, we set a threshold $\epsilon$ to constrains the minimal average anomaly probability in subgraph $\hat{G}_i$ and only the group in the subgraph that satisfies $\hat{S}_i^k \ge \epsilon$ will be extracted. Besides, a minimal extracted node number $\delta$ is adopted to ensure the extracted subgraph to consist of a potential group.

\subsection{Feature Deviation based GFM}
Different from node representation learning strategies in conventional graph tasks, anomalies in groups are subgraph deviations, which is hard to detect without group analysis of feature deviations. In order to capture the slight difference between connected nodes to conduct anomaly information aggregation, we propose a lightweight feature deviation-based graph foundation model based on GCN~\cite{KipfW17}. For $G_i$, the node embedding is designed as:
\begin{equation}
    H_i^{(L)}=f_\Theta(H_i^{(L-1)}, E_i),
\end{equation}
\begin{equation}
    h_u^{(L)} = \sigma [\sum_{r \in \mathcal{R}} \sum_{v \in \mathcal{N}_r(u)} W_r^{(L-1)}h_v^{(L-1)}+W_\text{self}^{(L-1)}h_u^{(L-1)}],
\end{equation}
\begin{equation}
    W_r = w_r \cdot \alpha_{uv} = w_r \cdot \phi(e^{-|x_u - x_v|}),
\end{equation}
where $\sigma$ is the ReLU activation function, $\mathcal{N}_r(u)$ indicates the neighbors of node $u \in G_i$ connected by edge type $r \in \mathcal{R}$, $h^{(0)}_u$ equals to the initial node features $x_u$, relational weight $w_r \in \mathbb{R}^{d \times d}$ is initialized by Gaussian distribution with normalization, and dot products with information aggregation weight between connected nodes $\alpha_{uv} \in \mathbb{R}$, which is calculated by a learnable two-layer Multilayer Perception $\phi$ with the exponential function $e^{(\cdot, \cdot)}$.

Once the final layer of node representation $H_i^{(L)}=\{h_v |v \in V_i\}$ is obtained, the average pooling is applied to obtain the embedding of subgraph $G_i$:
\begin{equation}
    h_{G_i} = \frac{1}{|V_i|} \sum_{v \in V_i} h_v.
\end{equation}

\subsection{Pretraining GFM with Dual-level Contrastive Learning}
Subgraphs containing group anomaly have group structural camouflage that can be similar to normal subgraphs. In order to discriminate the group anomaly patterns in those subgraphs, we select the subgraph $G_i$ that can extract a smaller subgraph $\hat{G}_i$ satisfying the threshold $\epsilon$ and node number $\delta$ mentioned above. Then, the subgraphs $(G_i, \hat{G}_i)$ can be seen as a positive pair, where $\hat{G}_i$ indicates high probability of group anomaly in the subgraph $G_i$.
For the subgraph $G_j$ that cannot extract a smaller subgraph $\hat{G}_j$ that meets the requirements of $\epsilon$ and $\delta$ will be considered as a negative subgraph. Following the InfoNCE loss~\cite{OORD18}, the estimated anomaly probability based subgraph-level contrastive learning is defined as:

\begin{equation}
    L_\text{sub} = - \log \frac{e^{(g_\text{s}(h_{G_i}, h_{\hat{G}_i}) / \tau)}}{e^{(g_\text{s}(h_{G_i}, h_{\hat{G}_i})/ \tau)} + \sum_j e^{(g_\text{s}(h_{G_i}, h_{G_j})/ \tau)}},
\end{equation}
where $h_G$ and $h_{\hat{G}}$ denote the subgraph representation of $G$ and $\hat{G}$, $g_\text{s}(\cdot, \cdot)$ is cosine similarity function, and $\tau$ is the temperature hyperparameter. 

Furthermore, it is also necessary to enhance the discrimination in group anomaly and normal parts inside the same subgraph $G_i$, so a node-level contrastive learning is employed to accurate estimate the group anomaly size in the subgraph. For nodes in subgraph $G_i$, connected nodes $u$ and $l$ with close estimated high anomaly probabilities are regarded as a positive pair, while non-connected nodes $u$ and $k$ with large difference in estimated anomaly probabilities are considered as a negative pair. The node-level contrast is defined in the following: 

\begin{equation}
    L_\text{node} = - \log \frac{\sum_l e^{(g_\text{s}(h_u, h_l) / \tau)}}{\sum_l e^{(g_\text{s}(h_u, h_l)/ \tau)} + \sum_k e^{(g_\text{s}(h_u, h_k)/ \tau)}}.
\end{equation}

Therefore, the graph foundation model $f_\Theta$ is pretrained on the dual-level contrastive learning tasks, with the subgraph-level task weight $\alpha$ to adjust the importance of the two types of contrastive learning tasks. The total pretraining loss is defined in the following, with the optimization for parameters $\Theta$ in graph foundation model and $\Phi$ in feature-based anomaly estimation module.

\begin{equation}
    \min_{\Theta_\text{pre}, \Phi_\text{pre}, } L_\text{pt} = \alpha \cdot L_\text{sub} + (1-\alpha) \cdot L_\text{node}.
\end{equation}

\subsection{Finetuning GFM for Group Anomaly Detection}
Different from induced subgraphs with target nodes in GFM for individual anomaly detection, the detection object is the whole subgraph in group anomaly detection, which is more challenging to accurately estimate members in anomalous groups across diverse size subgraphs. Under the conventional assumption of uniform node weight, large subgraphs exert disproportionate influence over the aggregated loss by virtue of their greater node counts, thereby inducing a representational bias toward the majority-class distribution, which undermines the discriminative ability in diverse group anomaly in the GFM. Therefore, we borrow the optimization strategy in imbalanced node classification task, and adjust node weights of group anomalies and non-anomalies in binary cross-entropy loss regarding to their corresponding proportions in the subgraph. 

Here, we assign $G_i^p=(V_i^p, D_i^p)$ as the group anomaly part, and $G_i^n=(V_i^n, D_i^n)$ as the normal part in $G_i$, where $V_i^p$ and $V_i^n$ denote node sets of group anomaly and non-anomaly, and $D_i^p$ and $D_i^n$ indicate the corresponding node degrees of group anomaly and non-anomaly. The weight of group anomaly $W_i^p$ and non-anomaly $W_i^n$ are defined as: 
\begin{equation}
\label{eq:w_pn}
    W_i^{\{p,n\}} = - W_i^G \cdot \log (\frac{|V_i^{\{p,n\}}|}{|V_i^p|+|V_i^n|}) \cdot D_i^{\{p,n\}},
\end{equation}
where $W_i^G=e^{1/{|V_i|}}$ denotes the subgraph weight determined by subgraph size. The last term in the above equation indicates that the majority part and nodes with high degrees can influence the structural and semantic characteristics of the subgraph.

Furthermore, the complexity of detecting group anomaly differs across graphs because of specific anomaly typologies and manifestation patterns. For two subgraphs, the instance characterized by a low anomaly ratio, in which anomalous nodes are sparsely scattered, exhibits markedly greater detection difficulty than its high-ratio counterpart, primarily because the elevated background noise and limited anomaly-specific context obscure discriminative signals. 

Therefore, we construct group anomaly context for each node, to upgrade the anomaly probability regarding to its anomaly-specific information in the subgraph. Firstly, the pretrained feature-based anomaly estimation module $f_{\Phi_\text{pre}}$ is finetuned to estimate the anomaly probability $S_i$ of each node in $G_i$ according to the node specific features in the downstream instances. Then, we compute the subgraph anomaly probability $S_i^G$ by averaging feature-based anomaly probabilities $S_i$:
\begin{equation}
    S_i^G = \text{Ave}(S_i)=\frac{1}{|V_i|} \sum_k S_i^k
\end{equation}

We filter the subgraph $G_i$ with $S_i^G$ larger than threshold $\epsilon$ to update anomaly probability based on group anomaly contexts. For node $u$ in $G_i$ whose anomaly probability is larger than $S_i^G$, we need ensure its group anomaly by similar nodes, and we select its top $K$ 1-hop neighbors $v \in \mathcal{N}(u)$ according to close anomaly probability and node degree, which construct the group anomaly context of node $u$:
\begin{equation}
    V_{\mathcal{N}(u)}^K = \text{Top K}[\{(1-|s_i^u-s_i^v|)\cdot(1-\frac{|d_i^u-d_i^v|}{d_i^u+d_i^v})\}],
\end{equation}

\begin{equation}
    h_u^\text{ctx}=\frac{1}{K}\sum_{m=1}^K h_u^m=\frac{1}{K}\sum_{m=1}^K f_\Theta(x_u^m).
\end{equation}
$x$ denotes the initial node features, and $h$ denotes node representations generated by the GFM $f_\Theta$.

For node $l$ whose anomaly probability is less than $S_i^G$, we need mitigate the its probability of anomaly camouflage by comparing with high potential anomalous nodes, so we select top $K$ 1-hop neighbors $v \in \mathcal{N}(l)$ but with large difference in anomaly probability and node degree to construct group anomaly context:

\begin{equation}
    V_{\mathcal{N}(l)}^K = \text{Top K}[\{|s_i^l-s_i^v|\cdot\frac{|d_i^l-d_i^v|}{d_i^l+d_i^v}\}].
\end{equation}

With potential group anomaly contexts, the anomaly probabilities can be updated by the bilinear group context-based anomaly estimation weight $W_\Psi$ with the sigmoid activation function $\sigma$:
\begin{equation}
    \hat{S}_i = f_\Psi (H_i^{(L)}, H_i^\text{ctx}) = \sigma (H_i^{(L)} W_\Psi H_i^\text{ctx}).
\end{equation}

Here, the $\hat{S}_i$ indicates the adjusted anomaly combining with group anomaly context in $G_i$. The final anomaly probability $\bar{S}_i=(S_i + \hat{S}_i)/2$ is averaged by the pretrained group knowledge $S_i$ and specific group anomaly information $\hat{S}_i$. Finally, the loss optimization of the GFM few-shot finetuning is defined as:
\begin{gather}
    L_\text{GA} = -\frac{1}{|\mathcal{G}_L^k|} \sum_i^{|\mathcal{G}_L^k|} [W_i^p \cdot y_i \cdot \log(\bar{S}_i) \nonumber + W_i^n \cdot (1-y_i) \cdot \log(1-\bar{S}_i)], \nonumber \\
    L_\text{FC} = ||\Theta - \Theta_\text{pre}||_2 + ||\Phi - \Phi_\text{pre}||_2 \nonumber, \\
    \min_{\Theta, \Psi} L_\text{ft} = L_\text{GA} + L_\text{FC}, 
\end{gather}
where $|\mathcal{G}_L^k|$ denotes the number of instances in the finetuning, and $W_p$ and $W_n$ are obtained from equation~(\ref{eq:w_pn}). The L2-regularization terms restrict changing magnitude in pretrained GFM parameters $\Theta_\text{pre}$ and group anomaly estimation $\Phi_\text{pre}$ throughout the few-shot finetuning phase.

\section{Experiments}
\subsection{Experimental Setup}
\textbf{Datasets.} We pretrain the GFM4GA on the real-world dataset provided by Weixin, and finetune it on the labeled group anomaly Weixin dataset and four public datasets with manually labeled individual anomalies: Weibo~\cite{DYJW020}, Facebook~\cite{XuHZDL22}, Amazon~\cite{DouL0DPY20}, and T-Finance~\cite{TangLGL22}. The  Weixin dataset contains 330K unlabeled subgraphs for pretraining, and has same feature dimension and edge type with its downstream dataset. As for the four public datasets, we sample subgraphs with connected anomalies from initial graphs to construct synthetic group anomaly datasets. Details of those datasets are listed as follows: 
\begin{itemize}
\item \textbf{Weixin} is a social network graph dataset, where each node represents a user, and there are multiple edge types between users. In this experiment, we are provided with 330k unlabeled subgraphs, and 598 manually labeled subgraphs with group anomalies.
\item \textbf{Weibo} is a public social network graph consisting of 8,405 nodes and 407,963 edges, containing 868 individual anomalies.
\item \textbf{Facebook} is a public social network graph with 4,039 nodes, 88,234 edges and 400 ground-truth individual anomalies. 
\item \textbf{Amazon} is a co-review network graph comprising 11,944 nodes, 175,608 edges and approximately 9.5\% individual anomalies.
\item \textbf{T-Finance} is a graph based on transaction network with 39,357 nodes, 21,222,543 edges and 4.58\% individual anomalies.
\end{itemize} 
The four synthetic group anomaly datasets derived from public individual anomaly datasets, which are constructed by randomly sampling 2-hop neighbors around the anomaly nodes and selecting subgraphs that contain at least three interconnected anomaly nodes. To avoid redundant information, we filter out similar subgraphs in the sampling set. The detailed information of those datasets for downstream finetuning and evaluation are demonstrated in Table~\ref{tab:data}. 
\begin{table}[htbp]
    \centering
    \caption{Detailed information of downstream datasets. 'Feat Dim' denotes feature dimension, 'Num' indicates the number of subgraphs, 'Ave Group' denotes average size of group anomaly in subgraphs, and 'Ave Subgraph Size' means the average size of subgraphs.}
    \begin{tabular}{ccccc}
    \toprule
    \small \textbf{Dataset} & \small \textbf{Feat Dim} & \small \textbf{Num} & \small \textbf{Ave Group} & \small \textbf{Ave Subgraph Size} \\
    \midrule
    \small Weixin & \small 768 & \small 598 & \small 28.35 & \small 98.68 \\
    \small Weibo & \small 400 & \small 338 & \small 7.98 & \small 36.25 \\
    \small Facebook & \small 576 & \small 256 & \small 3.04 & \small 10.18 \\
    \small Amazon & \small 25 & \small 384 & \small 5.76 & \small 19.57 \\
    \small T-Finance & \small 10 & \small 466 & \small 8.98 & \small 25.43 \\
    \bottomrule
    \end{tabular}
    \label{tab:data}
\end{table} \\
\textbf{Baselines.} We evaluate our proposed pipeline by comparing it with three categories of methods. Firstly, we select classical graph neural networks including GCN~\cite{KipfW17} and GAT~\cite{VelickovicCCRLB18} as the basic compared methods in group anomaly detection. Secondly, we employ several advanced graph anomaly detection methods. They include geneal GAD: (1) SL-GAD~\cite{ZhengJLCPC21}: a self-supervised graph anomaly detection method containing generative attribute regression and multi-view contrastive learning tasks. (2) DCL-GFD~\cite{YuLCH0W25}: a dynamic neighborhood modeling via node-subgraph contrastive learning for graph-based fraud detection. (3) ComGA~\cite{Luo22}: a community-aware attributed graph anomaly detection framework via feature and structure reconstructions. Lastly, several GFM-based GAD methods are chosen: (4) GUDI~\cite{Li024}: a few-shot-oriented graph anomaly detection approach with self-supervised pretraining across graph domains. (5) ARC~\cite{LZCZP24}: an "one-for-all" GFM-based pipeline to detect anomalies across various graph datasets via in-context learning. (6) UNPrompt~\cite{Niu24}: a zero-shot generalist GAD approach with latent node attribute as anomaly measure and neighbor prompt learning. (7) AnomalyGFM~\cite{Qiao25}: a graph foundation model that supports zero-shot inference and few-shot prompt tuning for GAD in diverse graph datasets. Following the previous approach~\cite{TangLGL22}, we employ AUROC and AUPRC to evaluate the performance across all datasets. \\
\textbf{Implementation Details.} The experiments are conducted on the environment of Python 3.9.0, PyTorch 1.13.1, and PyG 2.6.1, and hardware configurations include two NVIDIA V100 GPUs with 32 GB GPU memory on a Ubuntu 18.04 server. It spends about 30 hours to pretrain 50 epochs with steady pretraining loss in distributed data parallel mechanism. We pretrain two pipelines of GFM4GA on Weixin dataset, the first one is designed for the similar Weixin dataset in the downstream task, and the other one treats all edges as one type for the four public datasets. For the public datasets in the downstream experiments, we apply PCA algorithm to reduce the feature dimension of Weibo and Facebook to satisfy the input dimension of pretrained GFM, and add zero padding to align the dimension for Amazon and T-Finance. As we focus on few-shot anomaly detection scenarios, we have sampled different k-shot samples for downstream finetuning task. In each shot, we repeat the sampling for 5 times using 5 different seeds to increase the sampling diversities. Those hyperparameters are presented in Table~\ref{tab:parameters}. The detection targets on Weixin and synthetic datasets are all nodes in the subgraphs across all baseline methods, and we implement those baselines based on their open-sourced codes or algorithm descriptions provided in the respective papers.
\begin{table}[htbp]
    \centering
    \caption{Main hyperparameters in GFM4GA.}
    \begin{tabular}{c|c}
    \toprule
    \small \textbf{Hyperparamter name} & \small \textbf{Value} \\
    \midrule
    \small Graph encoder layers & \small 2 \\
    \small Contrastive temperature $\tau$ & \small 0.8 \\
    \small Subgraph average anomalies threshold $\epsilon$ & \small 0.4 \\
    \small Minimum members of group anomaly $\delta$ & \small 3 \\
    \small Subgraph-level contrastive weight $\alpha$ & \small 0.7 \\
    \small Input and hidden dimension for Weixin dataset & \small 768, 1024 \\
    \small Input and hidden dimension for public datasets & \small 64, 32 \\
    \small Finetuning shot numbers and interval & \small [10, 100], 10 \\
    \small Random seeds & \small [1, 2, 3, 4, 5] \\
    \bottomrule
    \end{tabular}
    \label{tab:parameters}
\end{table} 
\subsection{Overall Performance}
We compare the performance between GFM4GA and baselines in few-shot group anomaly detection, and the overall AUROC results of 10-shot finetuning are presented in Table~\ref{tab:auroc}, and AUPRC results are demonstrated in Table~\ref{tab:auprc}. The proposed GFM4GA achieves an average improvement of 2.85\% in AUROC and 2.55\% in AUPRC compared to the respective second-best results across those datasets, which indicates the dual-level contrastive learning based on anomaly estimation can better capture the group anomaly. The extracted smaller subgraph with a high anomaly score highlights dense abnormal connections within the original subgraph, and this augmentation facilitates more accurate localization of the group anomaly. It is worth noting that finetuning the pretrained pipeline in few-shot scenarios still achieves significantly better performance than prompt tuning, highlighting the limited effectiveness of graph prompt tuning strategies for group anomaly detection. Furthermore, the Weixin dataset features a larger average group size, offering richer contextual information for anomalies, whereas the synthetic group anomaly datasets have smaller group sizes, making them more akin to combinations of individual anomalies. As a result, ComGA achieves the second-best performance on the Weixin dataset, while GFMs designed for individual anomalies outperform other baselines on the synthetic group anomaly datasets.

\begin{table*}[htbp]
    \centering
    \caption{AUROC(\%) performance of group anomaly detection in 10-shot settings, reported with mean and standard deviation. Weibo, Facebook, Amazon and T-Finance are finetuned on the pipeline pretrained from Weixin. Bold and underline demonstrate the best and second best results. The 'Ave Rank' column displays the average AUROC ranking across datasets, and the last row indicates the improvement on each dataset and average improvement across datasets of GFM4GA over the second best results.}
    \label{tab:auroc}
    \renewcommand{\arraystretch}{1.2}
    \begin{tabular}{c|cccccc}
    \toprule
    \textbf{Dataset} & \textbf{Weixin} & \textbf{Weibo} & \textbf{Facebook} & \textbf{Amazon} & \textbf{T-Finance} & \textbf{Ave Rank} \\
    \midrule
    GCN & 56.23$\pm$1.81 & 55.57$\pm$2.63 & 64.28$\pm$1.84 & 52.35$\pm$1.82 & 52.28$\pm$1.91 & 8.2 \\
    GAT & 58.71$\pm$2.18 & 56.72$\pm$2.15 & 67.52$\pm$1.55 & 53.87$\pm$2.54 & 53.94$\pm$1.83 & 6.8 \\
    \hline
    SL-GAD & 68.19$\pm$1.98 & 60.45$\pm$2.38 & 66.43$\pm$1.89 & 55.54$\pm$2.31 & 55.68$\pm$3.86 & 6.0 \\
    DCL-GFD & 76.85$\pm$3.09 & 62.21$\pm$1.87 & 69.22$\pm$1.14 & 56.46$\pm$1.68 & 56.49$\pm$2.32 & 4.0 \\
    ComGA & \underline{77.61$\pm$2.56} & 61.02$\pm$2.56 & 68.65$\pm$1.57 & 55.29$\pm$1.93 & \underline{58.38$\pm$1.87} & 4.0 \\
    \hline
    GUDI & 73.64$\pm$2.87 & \underline{63.25$\pm$0.37} & 69.84$\pm$3.29 & \underline{58.78$\pm$1.03} & 58.02$\pm$1.85 & \underline{3.0} \\
    ARC & 74.56$\pm$2.07 & 62.34$\pm$1.89 & \underline{71.24$\pm$1.62} & 40.26$\pm$1.97 & 44.61$\pm$2.63 & 3.2 \\
    UNPrompt & 52.62$\pm$2.60 & 42.04$\pm$1.79 & 46.73$\pm$2.63 & 40.26$\pm$1.97 & 44.61$\pm$2.63 & 9.8 \\
    AnomalyGFM & 56.67$\pm$2.73 & 44.37$\pm$2.98 & 49.28$\pm$2.69 &  41.58$\pm$2.86 & 42.36$\pm$2.56 & 9.0 \\
    \hline
    \textbf{GFM4GA(Ours)} & \textbf{80.51$\pm$1.23} & \textbf{65.04$\pm$1.28} & \textbf{73.73$\pm$1.42} & \textbf{62.21$\pm$2.56} & \textbf{62.01$\pm$0.76} & 1.0 \\
    \hline
    Improvement & +2.90\% & +1.79\% &  +2.49\% &  +3.43\% & +3.63\% & +2.85\% (Ave Imp.) \\
    \bottomrule
    \end{tabular}
\end{table*}

\begin{figure}[htbp]
    \centering
    \includegraphics[width=0.8\linewidth]{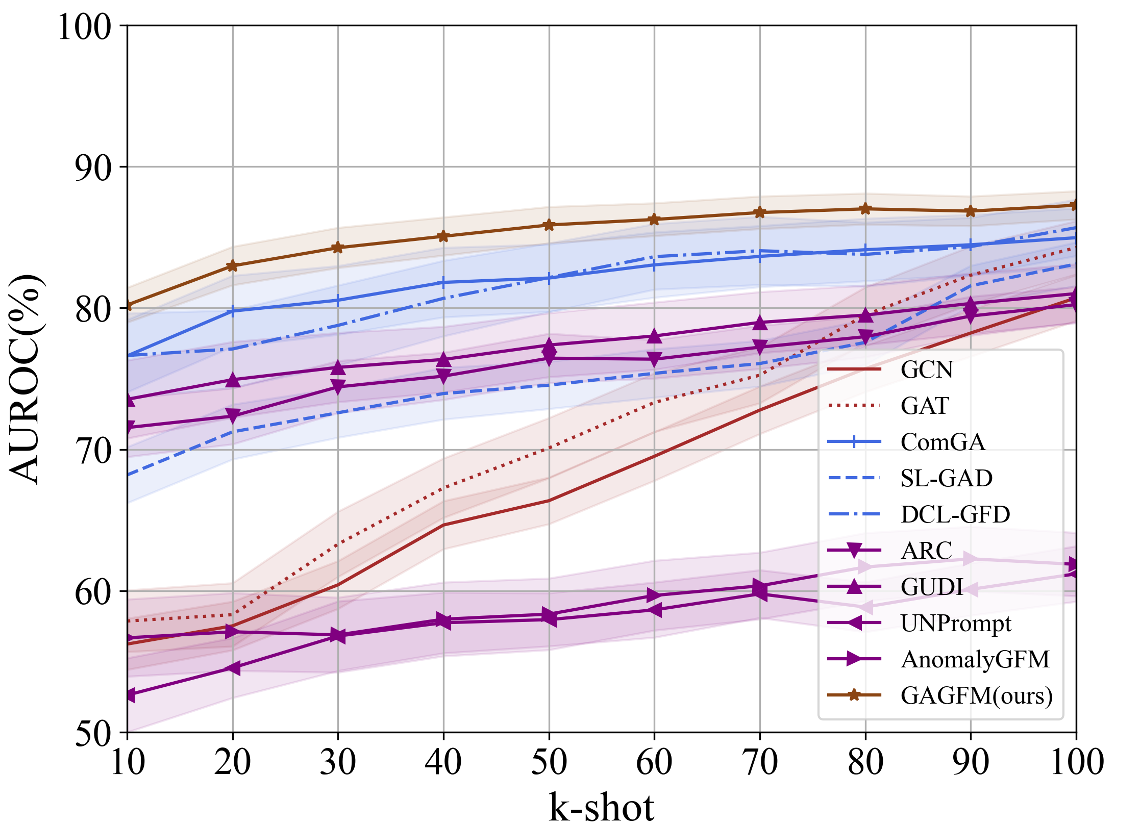}
    \caption{AUROC performance of k-shot on Weixin dataset, and shaded areas around lines denote standard deviations.}
    \label{fig:kshot}
\end{figure}

Besides, we vary the k-shot number from 10 to 100 on Weixin dataset to explore the trend of few-shot finetuning performance. It is also worth emphasizing that the proposed GFM4GA can be much better than baselines when the finetuning subgraphs are below 50 shot, which demonstrates the strength of the proposed pipeline in severe few-shot scenarios when detecting group anomaly. The graph prompt–based GFM methods, UNPrompt and AnomalyGFM, exhibit suboptimal performance in few-shot group anomaly detection scenarios, which can be attributed to the limited discriminative information contained in the upstream pretraining data, and the unstable prompt initialization and tuning strategy. Compared to other methods, GFM4GA exhibits a narrower standard deviation range, demonstrating more consistent and stable results.

\begin{figure}[H]
    \centering
    \includegraphics[width=0.76\linewidth]{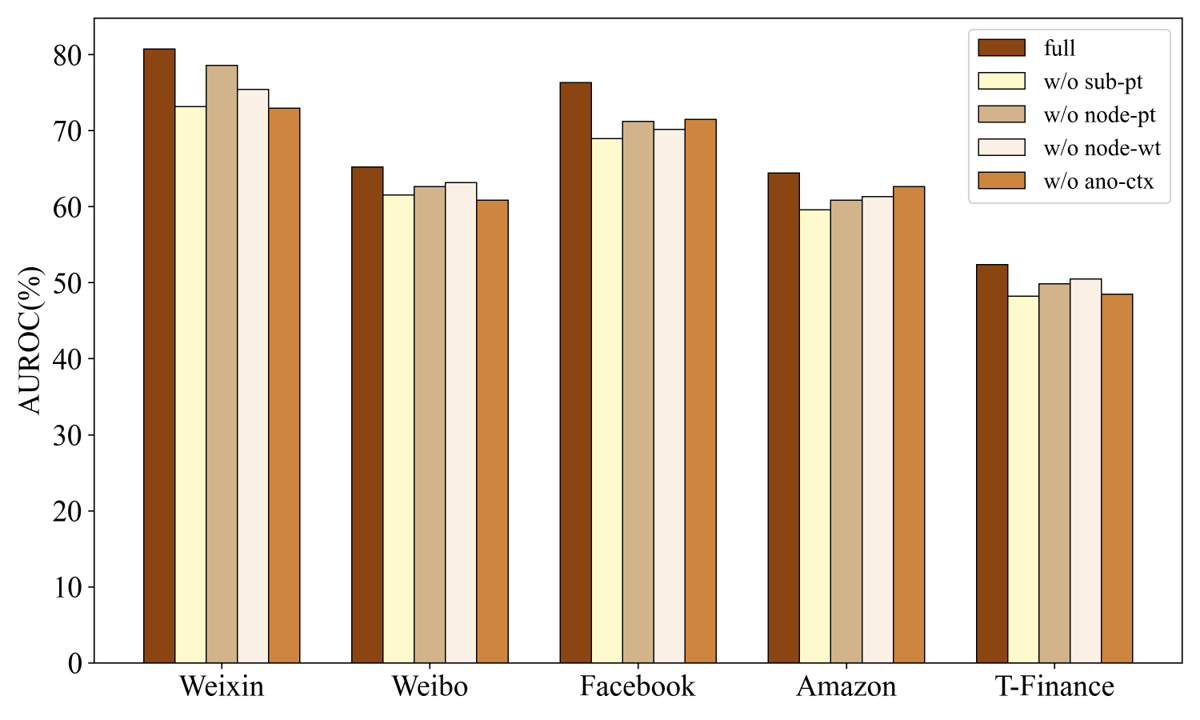}
    \caption{AUROC results of ablation studies across datasets.}
    \label{fig:ablation}
\end{figure}

\begin{table*}[htbp]
    \centering
    \caption{AUPRC(\%) performance of group anomaly detection in 10-shot settings, reported with mean and standard deviation. Detailed meaning of each column is similar to the table in the above, except for the 'Ave Rank' column displays the average AUPRC ranking across datasets.}
    \renewcommand{\arraystretch}{1.2}
    \begin{tabular}{c|cccccc}
    \toprule
    \textbf{Dataset} & \textbf{Weixin} & \textbf{Weibo} & \textbf{Facebook} & \textbf{Amazon} & \textbf{T-Finance} & \textbf{Ave Rank} \\
    \midrule
    GCN & 45.55$\pm$1.62 & 43.63$\pm$1.74 & 52.15$\pm$2.03 & 42.03$\pm$1.66 & 43.21$\pm$3.68 & 8.6 \\
    GAT & 45.92$\pm$1.37 & 42.65$\pm$1.87 & 54.24$\pm$0.98 & 42.65$\pm$2.68 & 44.61$\pm$2.47 & 8.4 \\
    \hline
    SL-GAD & 57.64$\pm$1.87 & 46.85$\pm$3.01 & 53.23$\pm$1.65 & 46.33$\pm$1.83 & 47.33$\pm$1.24 & 5.8 \\
    DCL-GFD & \underline{64.32$\pm$1.25} & 48.18$\pm$1.67 & 55.88$\pm$2.73 & 47.05$\pm$2.76 & 48.67$\pm$2.25 & 3.6 \\
    ComGA & 63.54$\pm$1.42 & 46.87$\pm$2.26 & 54.46$\pm$1.88 & 45.24$\pm$3.17 & 49.67$\pm$1.25 & 4.4 \\
    \hline
    GUDI & 60.12$\pm$1.77 & \underline{49.26$\pm$1.35} & 54.12$\pm$1.34 & 49.34$\pm$1.24 & \underline{52.17$\pm$2.71} & 3.4 \\
    ARC & 62.36$\pm$1.33 & 48.92$\pm$1.76 & \underline{56.36$\pm$1.53} & \underline{50.12$\pm$1.24} & 50.97$\pm$1.80 & 2.8 \\
    UNPrompt & 42.27$\pm$2.49 & 33.74$\pm$2.49 & 38.36$\pm$1.85 & 36.58$\pm$2.01 & 44.47$\pm$2.67 & 9.6 \\
    AnomalyGFM & 46.21$\pm$2.86 & 38.59$\pm$2.64 & 41.51$\pm$2.77 & 37.24$\pm$2.69 & 43.34$\pm$2.24 & 8.6 \\
    \textbf{GFM4GA(Ours)} & \textbf{66.98$\pm$2.17} & \textbf{51.03$\pm$1.34} & \textbf{56.71$\pm$1.54} & \textbf{53.65$\pm$2.79} & \textbf{56.61$\pm$1.14} & \textbf{1.0} \\
    \midrule
    Improvement & +2.66\% & +1.77\% & +0.35\% & +3.53\% & +4.44\% & +2.55\% (Ave Imp.) \\
    \bottomrule
    \end{tabular}
    \label{tab:auprc}
\end{table*}

\begin{figure*}[htbp]
    \centering
    \begin{subfigure}[t]{0.3\textwidth}
        \centering
        \includegraphics[width=\linewidth]{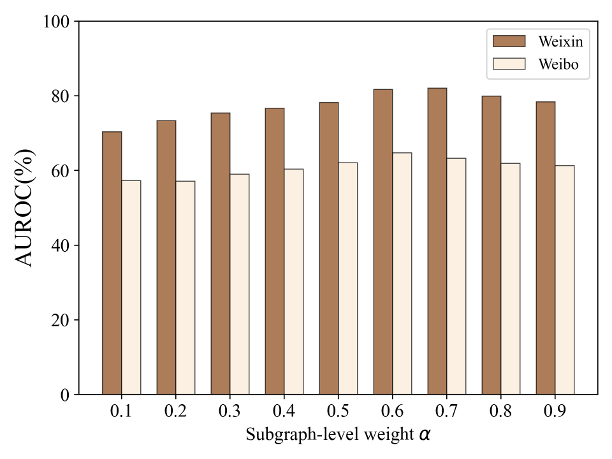}
        \caption{AUROC performance of 10-shot finetuning with different $\alpha$ on Weixin and Weibo datasets.}
        \label{fig:alpha}
    \end{subfigure}
    \begin{subfigure}[t]{0.3\textwidth}
        \centering
        \includegraphics[width=\linewidth]{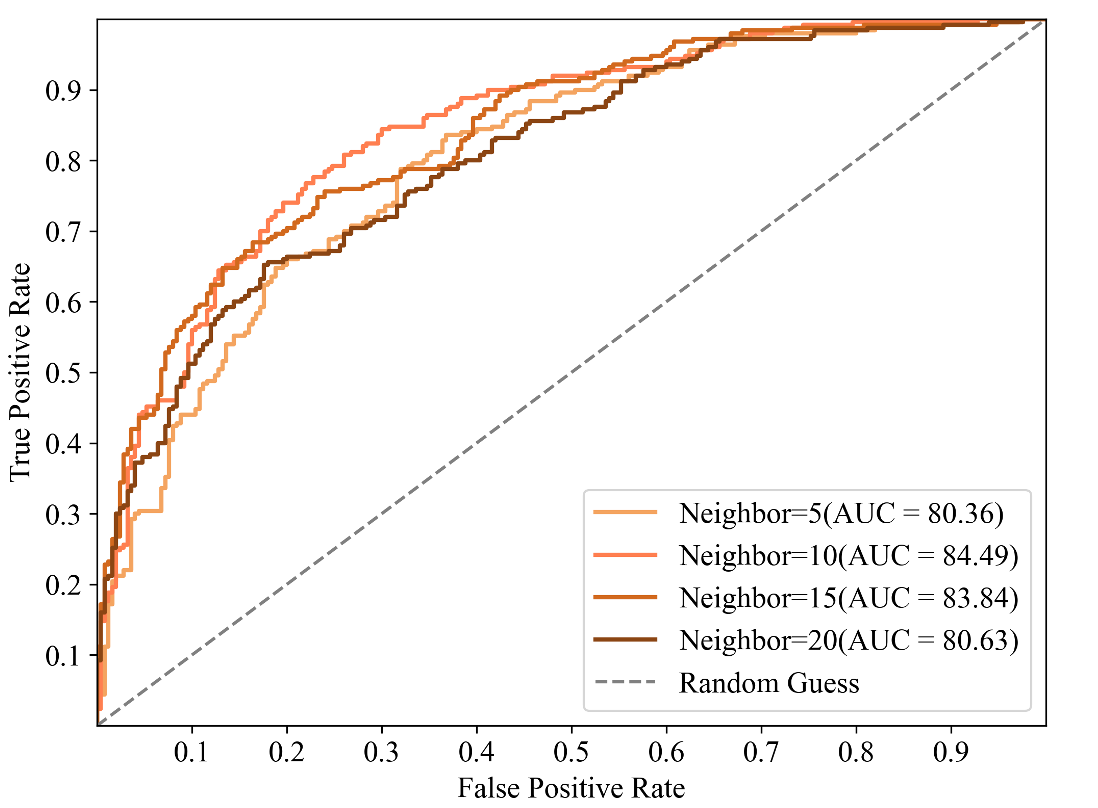}
        \caption{ROC curves of group anomaly contexts constructed from top $K$ neighbors on Weixin dataset.}
        \label{fig:param1}
    \end{subfigure}
    \begin{subfigure}[t]{0.3\textwidth}
        \centering
        \includegraphics[width=\linewidth]{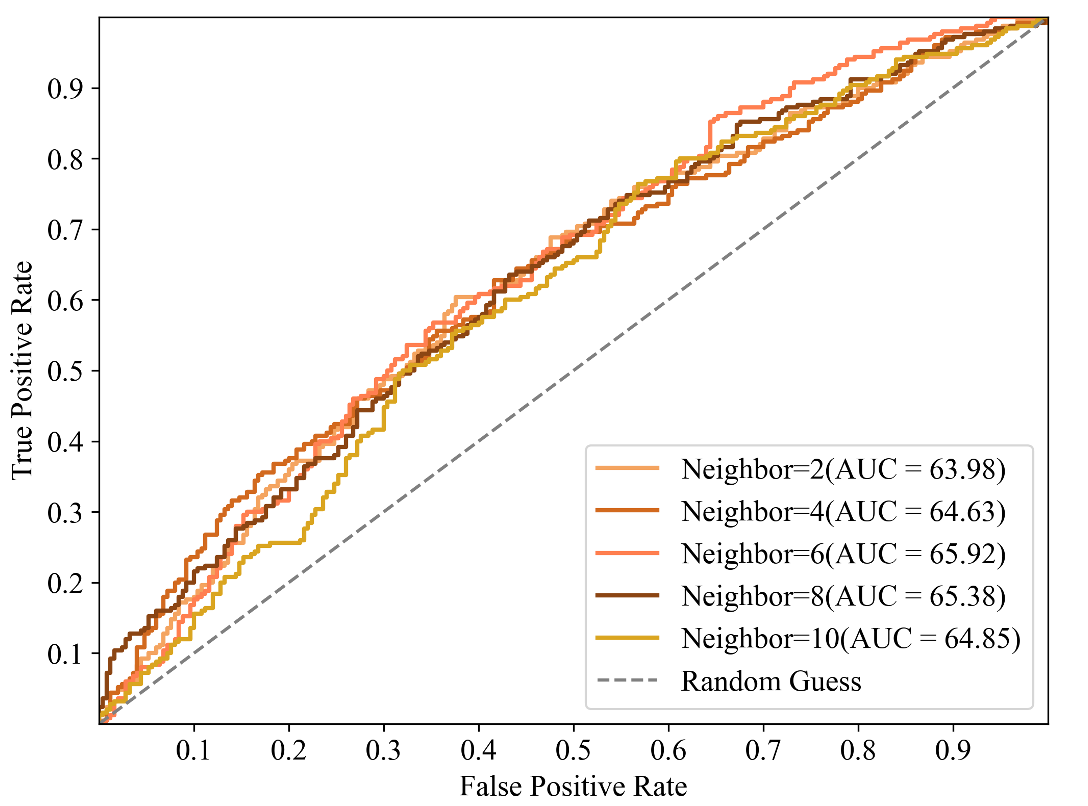}
        \caption{ROC curves of group anomaly contexts constructed from top $K$ neighbors on Weibo dataset.}
        \label{fig:param2}
    \end{subfigure}
    \caption{Experimental results of parameter sensitivity.}
    \label{fig:all}
\end{figure*}
\subsection{Ablation Studies}
To examine the contribution of each module in GFM4GA, we conduct ablation studies, removing each element while keep the rest of the pipeline intact. The elements include (1) subgraph-level contrastive learning (sub-pt), (2) node-level contrastive learning (node-pt), (3) adjusting weight in few-shot finetuning (node-wt), and (4) anomaly group context in few-shot finetuning (ano-ctx). The AUROC results are depicted in Figure~\ref{fig:ablation}, which shows the performance declines after the removal of each element. It is worth noting that the largest decrease occurs without subgraph contrastive learning in the pretraining phase, which addresses the essence of group anomaly pattern learned in subgraphs contrast. Besides, anomaly contexts also play a crucial role in group anomaly detection, particularly in datasets such as Weixin and Weibo, where group anomalies constitute only a small proportion of each subgraph. This observation suggests that a lower ratio of group anomalies within subgraphs requires richer group-level contextual information for effective detection.
\subsection{Parameter Sensitivity}
We report AUROC results on Weixin and Weibo datasets to examine the sensitivity of two key hyperparameters: the subgraph-level contrastive weight $\alpha$, and the number of top $K$ neighbors to construct group contexts. \\
\textbf{Subgraph-level contrastive weight $\alpha$.} In the pretraining phase, the subgraph-level and node-level contrastive weights are defined as $\alpha=0.5$ and $1-\alpha=0.5$ by default. Then, we adjust the $\alpha$ from 0.1 to 0.9, to investigate the importance of the two contrastive learning in downstream 10-shot finetuning. The performance, shown in Figure~\ref{fig:alpha}, highlights the critical importance of subgraph-level contrast in group anomaly detection. \\
\textbf{Top $K$ neighbors to construct group context.} In the finetuning phase, the number of neighbors to construct group anomaly contexts is varied from 5 to 20 with a step size of 5 on Weixin dataset. The corresponding results are reported in Figure~\ref{fig:param1}. It can be seen that when $K$ equals to 10, the AUC achieves the overall best result. Similarly, we conduct similar experiment on the Weibo dataset, varying the number of top $K$ neighbors from 2 to 10 in the step of 2, where the AUC peaks at 6 in Figure~\ref{fig:param2}. Compared with the Weixin dataset, the changes in AUC performance are smaller. This phenomenon can be attributed to two potential factors: first, the pretrained GFM may impose constraints on cross-domain generalization; second, the Weibo dataset may be limited in graph scale and the richness of contextual information. We posit that the effectiveness of selecting neighbors as group anomaly context is inherently influenced by the statistical properties of each dataset.

\section{Time Analysis}
In this section, we provide a detailed analysis of the time complexity of GFM4GA. First, the group anomaly estimation module which integrates PCA and a MLP, and requires a computation cost of $O_1 = \mathcal{O}(|\mathcal{G}_U|N_\mathcal{G}(d^2+d))$, where $|\mathcal{G}_U|$ is the number of unlabeled subgraphs used for pretraining, $N_\mathcal{G}$ is the average nodes per graph in the dataset $\mathcal{G}$, and $d$ denotes the feature dimensionality. The complexity of dual-level contrastive learning based on a feature deviation based GCN takes $O_2 = \mathcal{O}(2L|\mathcal{G}_U|(KN_\mathcal{G} + KE_\mathcal{G} + BN_\mathcal{G}^a + BN_\mathcal{G}^a)d])$, where $L$ is the layer number, $E_\mathcal{G}$ denotes the average edges of the dataset, $K$ is the average number of negative pair, $B$ is the batch size, and $N_\mathcal{G}^a$ and $N_\mathcal{G}^a$ indicate the average nodes and edge potential group anomaly in the dataset. Thus, the overall complexity of pretraining GFM4GA is $O_\text{pt} = O_1 + O_2$. During the finetuning phase, additional operations are introduced, including anomaly estimation via group-level contextual information and L2-regularization on the pretrained parameters. The corresponding computational complexity is given by $O_\text{ft} = O_\text{pt} + \mathcal{O}(|\mathcal{G}_L^k|N_\mathcal{G}d + P)$, where $|\mathcal{G}_L^k|$ is the number of labeled subgraphs for k-shot finetuning, and $P$ denotes the constant computation introduced by the L2 regularization term.

\section{Conclusion}
We address the issues in GFM for group anomaly detection. Specifically, we extract potential group based on feature-based anomaly probability, and construct subgraph-level and node-level contrastive learning between the initial and extracted subgraphs to capture group structural pattern and feature discrimination via feature deviation based GFM. Since the detection objects are subgraphs with diverse sizes and group anomalies in the downstream few-shot finetuning phase, we adjust node weights according to group anomaly proportions of the node in the subgraph. We also construct group anomaly context based on local neighbors, to provide discriminative information to ensure the accurate estimation of group anomaly. Extensive experiments across datasets demonstrate the proposed GFM4GA performs better on few-shot group anomaly detection than those baselines in AUROC and AUPRC metrics. Future research will investigate the effectiveness of LLMs and LLM-based GFMs for group anomaly detection when textual information is available. Additionally, the reasoning capabilities of LLMs can be explored in zero-shot scenarios of group anomaly detection.

\bibliographystyle{ACM-Reference-Format}
\bibliography{GFM4GA_arXiv_reference}

\end{document}